\documentclass{article}

\usepackage{arxiv}

\usepackage[utf8]{inputenc} % allow utf-8 input
\usepackage[T1]{fontenc}    % use 8-bit T1 fonts
\usepackage{url}            % simple URL typesetting
\usepackage{booktabs}       % professional-quality tables
\usepackage{amsfonts}       % blackboard math symbols
\usepackage{nicefrac}       % compact symbols for 1/2, etc.
\usepackage{microtype}      % microtypography
\usepackage{lipsum}
\usepackage{graphicx}
\usepackage{amsmath,amsfonts}
\usepackage{algorithmic}
\usepackage{algorithm}
\usepackage{array}
\usepackage[caption=false,font=normalsize,labelfont=sf,textfont=sf]{subfig}
\usepackage{textcomp}
\usepackage{stfloats}
\usepackage{verbatim}
\usepackage{cite}
\usepackage{booktabs}
\usepackage{algorithmic}
\usepackage{algorithm}
\usepackage{multirow}
\graphicspath{ {./images/} }

\title{Exploiting Multiple Representations: 3D Face Biometrics Fusion with Application to Surveillance}

\author{Simone Maurizio La Cava$^1$, Roberto Casula$^1$, Sara Concas$^1$, Giulia Orrù$^1$, Ruben Tolosana$^2$, \\ \textbf{Martin Drahansky$^3$, 
Julian Fierrez$^2$, Gian Luca Marcialis$^1$}\\
$^1$University of Cagliari, Cagliari, Italy \\e-mail: \{simonem.lac, roberto.casula, sara.concas90c, giulia.orru, gianluca.marcialis\}@unica.it
\\
$^2$Autonomous University of Madrid, Madrid, Spain \\\texttt{e-mail: \{ruben.tolosana,julian.fierrez\}@uam.es}
\\
$^3$Police Academy of the Czech Republic in Prague, Czech Republic \\
email: drahansky@polac.cz
}

\begin{document}
\maketitle
\begin{abstract}
3D face reconstruction (3DFR) algorithms are based on specific assumptions tailored to the limits and characteristics of the different application scenarios. In this study, we investigate how multiple state-of-the-art 3DFR algorithms can be used to generate a better representation of subjects, with the final goal of improving the performance of face recognition systems in challenging uncontrolled scenarios. We also explore how different parametric and non-parametric score-level fusion methods can exploit the unique strengths of multiple 3DFR algorithms to enhance biometric recognition robustness. With this goal, we propose a comprehensive analysis of several face recognition systems across diverse conditions, such as varying distances and camera setups, intra-dataset and cross-dataset, to assess the robustness of the proposed ensemble method. % in a surveillance context.
The results demonstrate that the distinct information provided by different 3DFR algorithms can alleviate the problem of generalizing over multiple application scenarios. 
In addition, the present study highlights the potential of advanced fusion strategies to enhance the reliability of 3DFR-based face recognition systems, providing the research community with key insights to exploit them in real-world applications effectively. Although the experiments are carried out in a specific face verification setup, our proposed fusion-based 3DFR methods may be applied to other tasks around face biometrics that are not strictly related to identity recognition.
\end{abstract}

% keywords can be removed

\section{Introduction}
\begin{figure}[]
\centering
\includegraphics[width=0.75\linewidth]{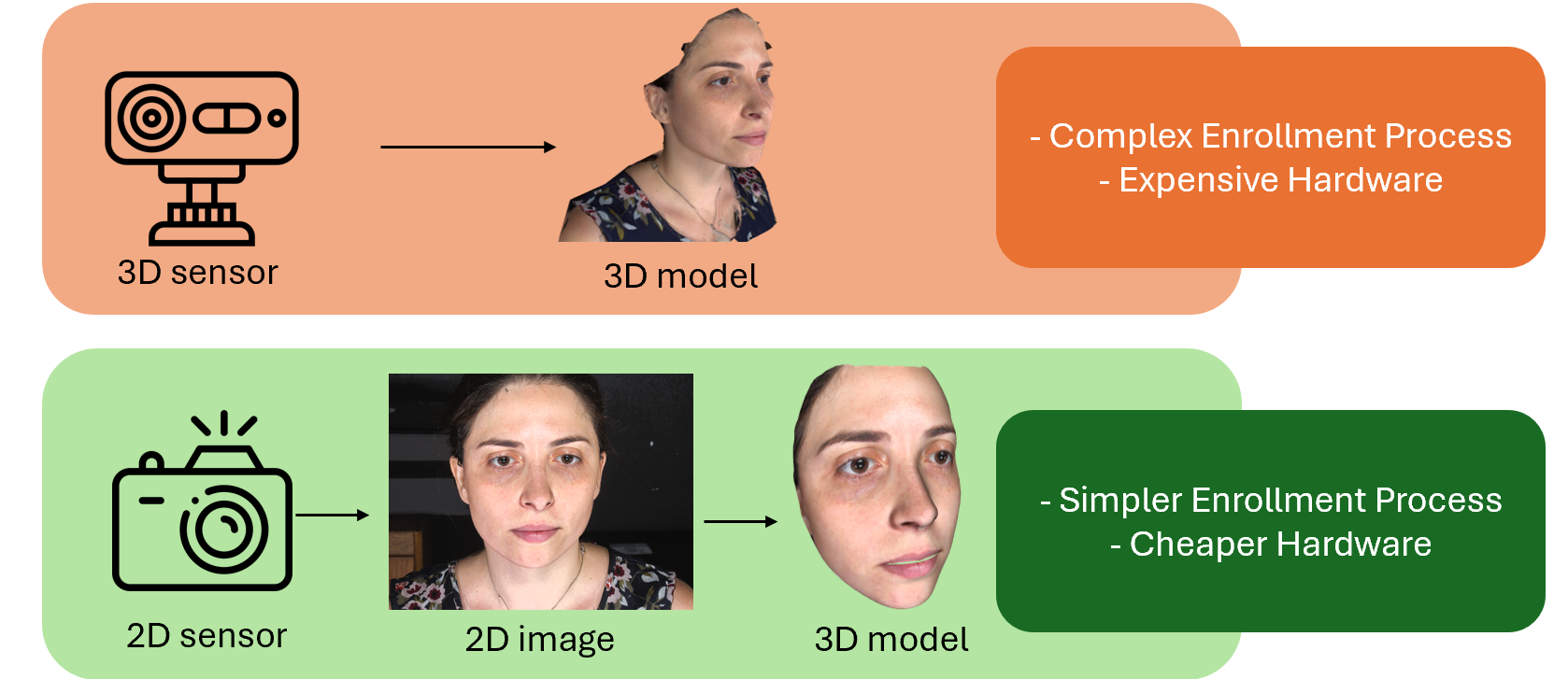}
\caption{Advantages of 3D face reconstruction from a 2D acquisition in comparison with 3D data acquisition.}
\label{fig:3DreconstructionExample}
\end{figure}

In recent years, the synthesis of facial images has received significant attention \cite{deandres2024frcsyn,shahreza2024sdfr,melzi2024frcsyn,syntheticreview2024}. This is especially relevant for 3D data, due to the robustness to adverse environmental factors such as poor lighting conditions and non-frontal facial poses \cite{la20223d,sun2022controllable,faceqvec22,quality22}. Furthermore, the literature has proven that better representations can be achieved in biometric recognition due to complementary information on shape and texture \cite{3Dstereo,ICPrecognition}. However, 3D data acquisition requires a more complex enrollment process and expensive hardware than standard 2D images, making it unsuitable in most application scenarios \cite{la20233d,uncalibrated}. For this reason, current facial recognition technology is still mostly based on 2D image acquisition, which is cheaper and more accessible.
Nevertheless, 3D face reconstruction (3DFR) algorithms from 2D images and videos can be a good solution to overcome the limitations of 2D data, combining the ease of 2D data acquisition with the robustness of 3D facial models \cite{la20223d}, as can be seen in Figure \ref{fig:3DreconstructionExample}. 

3DFR algorithms have proven particularly suitable in challenging scenarios where a face acquired under different and unconstrained conditions, even with arbitrary poses and distances, is paired with an identity using a good-quality reference image \cite{la20233d}. However, it is important to note that each 3DFR algorithm is usually designed for a specific scenario and may not necessarily generalize well to other contexts \cite{uncalibrated}. For this reason, the researcher ``maximizes'' specific features, such as fine details or individual facial components, that are useful in the particular application scenario \cite{chai2022realy}. %As a result, different 3DFR algorithms have different strengths and limitations, resulting in their varying effectiveness in individual use cases.

Based on these observations, this study aims to advance in this direction: \textit{can we improve face recognition performance in uncontrolled scenarios by combining complementary information provided by different 3DFR algorithms?} Accordingly, we propose using different 3DFR algorithms to individually support the training of multiple classifiers, finally adopting a score-level fusion rule to moderate their matching scores \cite{concas2022analysis}. The explored task is referred to as personal verification, determining whether the subject's identity in the reference data matches the one in the test data. Without loss of generality, we have chosen to assess the reliability of the proposed approach to surveillance \cite{proencca2018trends} as it represents a very challenging scenario \cite{tome2015combination,sosa2018}.
To sum up, the main contributions of the present study are the following:
\begin{itemize}
\item The analysis of the complementary information provided by different 3DFR algorithms, not strictly proposed for verification purposes, to various deep-learning systems with significantly different architectural complexities.
\item The exploitation of such complementarity to enhance intra- and cross-dataset verification performance in the particular case of surveillance scenarios; these are characterized by data acquired in known and unknown environmental conditions and camera-distance settings.
\item The exploration of the effectiveness, advantages, and disadvantages of non-parametric and parametric score-level fusion methods to take advantage of the complementary information.
\item Guidelines to aid technologists and operators in designing and implementing biometric systems based on the proposed ensemble-based approach according to the specific application scenario.
\end{itemize}

A preliminary version of this paper was published in \cite{la2024exploring}. We significantly improve that in the following aspects: (i) we include further state-of-the-art deep learning systems, typically employed in surveillance scenarios, with significantly different architectural complexities, in the analysis of the complementary information provided by distinct 3DFR algorithms, (ii) we integrate an additional 3DFR algorithm in the investigation, (iii) we extend the experiments to parametric score-level fusion methods, therefore analyzing their capability in taking advantage of 3DFR information to enhance face verification in surveillance, (iv) we perform a more detailed intra-dataset evaluation of the proposed approach to analyze the robustness of the system under challenging conditions, \textit{i.e.}, when dealing with data acquired in a different acquisition setting than the one considered for the system design, in terms of acquisition distance and surveillance cameras, (v) we assess its generalisation ability cross-dataset scenarios, (vi) we provide key insights to guide the research community to the most appropriate ensemble-based setup according to the specific application scenario.

The rest of the paper is organized as follows. Section \ref{sec:relatedwork} discusses the state of the art of 3DFR in surveillance and fusion methods in face recognition. Section \ref{sec:proposed} describes the proposed approach based on 3DFR algorithms and fusion methods. Section \ref{sec:framework} reports the experimental protocol. The experimental results are reported in Section \ref{sec:results}. Finally, a set of key insights and guidelines are provided in Section \ref{sec:guidelines} and conclusions are drawn in Section \ref{sec:conclusions}.

\section{Related Works} \label{sec:relatedwork}

\subsection{Enhancing Face Recognition through 3DFR Algorithms}
\label{subsec:personalRecognition}

3DFR permits biometric recognition systems to be more robust against intra-class variations, especially pose variations, since those systems often encounter probe images representing non-frontal faces due to unconstrained acquisition \cite{tome2013,sosa2018,HU201746,la20223d,quality22}. 
However, the extent of performance improvement when transitioning from 2D to 3DFR methods heavily relies on the specific approaches employed for integrating 3DFR into the face recognition process.
In particular, these approaches can be categorized into model- or view-based methods \cite{HanJain2012}. 

Model-based approaches involve synthesizing frontal faces from images with non-frontal views. These frontalized faces are then compared with the gallery images to identify the subject \cite{HanJain2012}.
This approach can introduce textural artifacts in the synthesized frontal images \cite{hassner2015effective,Cao2020,RotateAndRender}, making it more suitable for face identification tasks rather than high-accuracy authentication \cite{HanJain2012}.

On the other hand, view-based approaches adapt the images containing frontal faces to non-frontal ones, generating lateral views from the 3D template. The 3D facial model can be projected into various 2D poses to enhance each subject's representation \cite{HanJain2012,Zhang2008,Liang2018,Liang2020}. 
Despite more computational and storage requirements, these approaches are preferred for verification tasks requiring high reliability \cite{HanJain2012}.

\subsection{3DFR from Uncalibrated Images}
\label{subsec:3dfrANDvsurveillance}

%Face recognition in surveillance scenarios must generally rely on uncalibrated images, presenting unique challenges due to the wide range of lighting, pose, and scale variations caused by environmental conditions and the subject's non-cooperativeness \cite{la20233d}.

3DFR involves estimating the pose of the head, the facial geometry, and texture from the available data. However, face recognition in uncontrolled scenarios must generally rely on uncalibrated images, making the reconstruction an inherently ill-posed problem because multiple 3D faces could generate the same 2D image \cite{uncalibrated}.
To address this issue, researchers have proposed incorporating prior knowledge to estimate the actual 3D geometry. This information can be integrated using three main methods: photometric stereo, statistical model fitting, and deep learning.

The first approach analyses a 3D template with photometric stereo techniques to estimate the facial surface model by the shading patterns under different lighting conditions. Although it captures fine details, photometric stereo typically requires multiple images, which can be an unsuitable constraint in certain applications \cite{3Dstereo,uncalibrated}. 

Statistical model fitting adapts a pre-constructed 3D facial model to the input images. The 3D Morphable Model (3DMM) is a popular statistical model that includes a shape model and, optionally, an albedo model, both built using Principal Component Analysis (PCA) on a set of 3D facial scans \cite{10.1145/3596711.3596730}. This approach adjusts the model to fit the input images, providing generally convincing results with lower computational complexity. However, it tends to focus on global facial characteristics and may lack fine detail \cite{uncalibrated}.

Similarly, the approach based on deep learning maps 2D images to 3D shapes through deep neural networks trained on 3D scans. Although this approach can produce detailed reconstructions even from single images, it requires large datasets of 3D scans to generate reliable 3D models \cite{uncalibrated}.

As each approach has its strengths and weaknesses, further discussed for various application scenarios in \cite{la20233d} and \cite{uncalibrated}, the complementary nature of these methods suggests that combining multiple 3DFR algorithms could enhance the reliability of the reconstructed 3D faces \cite{geng2020towards,rowan2024n}. For instance, previous studies highlight that it is possible to reconstruct highly detailed 3D faces even with a single image by combining the prior knowledge of the global facial shape encoded in the 3DMM and refining it through a photometric or a deep learning approach \cite{la20233d}.
Starting from these assumptions, in this article, we analyze that complementarity in a surveillance scenario, proposing an approach based on fusion techniques to leverage the different strengths of distinct 3DFR algorithms in a face verification task, similarly as it has been very successfully applied to other tasks related to the face \cite{concas2022analysis,concas2024quality} and other biometrics like fingerprint \cite{finger05} or speaker recognition \cite{speaker06}.

\subsection{Ensemble Methods for Face Recognition} \label{sec:fusion}

Ensemble methods and uni-modal or multi-modal fusion approaches are widely used in pattern recognition to improve generalization and handle intra- and inter-class variations \cite{fusion18-2,panzino2024evaluating}. In biometrics, fusion can be performed at various levels of the classification system, including at the sensor, feature, score, and decision levels \cite{fusion18-2}.
Among others, score-level fusion is particularly advantageous as it leverages the complementarity of different methods without significantly increasing system complexity, as in the case of sensor- and feature-level fusion, and provides more nuanced information than decision-level fusion \cite{wang2022survey, marcialis2004fusion}. 

Numerous studies have supported the effectiveness of score-level fusion in face recognition, dating back to early attempts in the field \cite{fusion18-1,fusion18-2,sim2014multimodal,punyani_evaluation_2017}. Based on previous research, we hypothesize that integrating multiple 3DFR algorithms through score-level fusion can enhance the robustness and accuracy of face recognition systems, particularly in challenging surveillance scenarios. 

\section{Proposed Method}
\label{sec:proposed}

\begin{figure*}
\centering
\includegraphics[width=\linewidth]{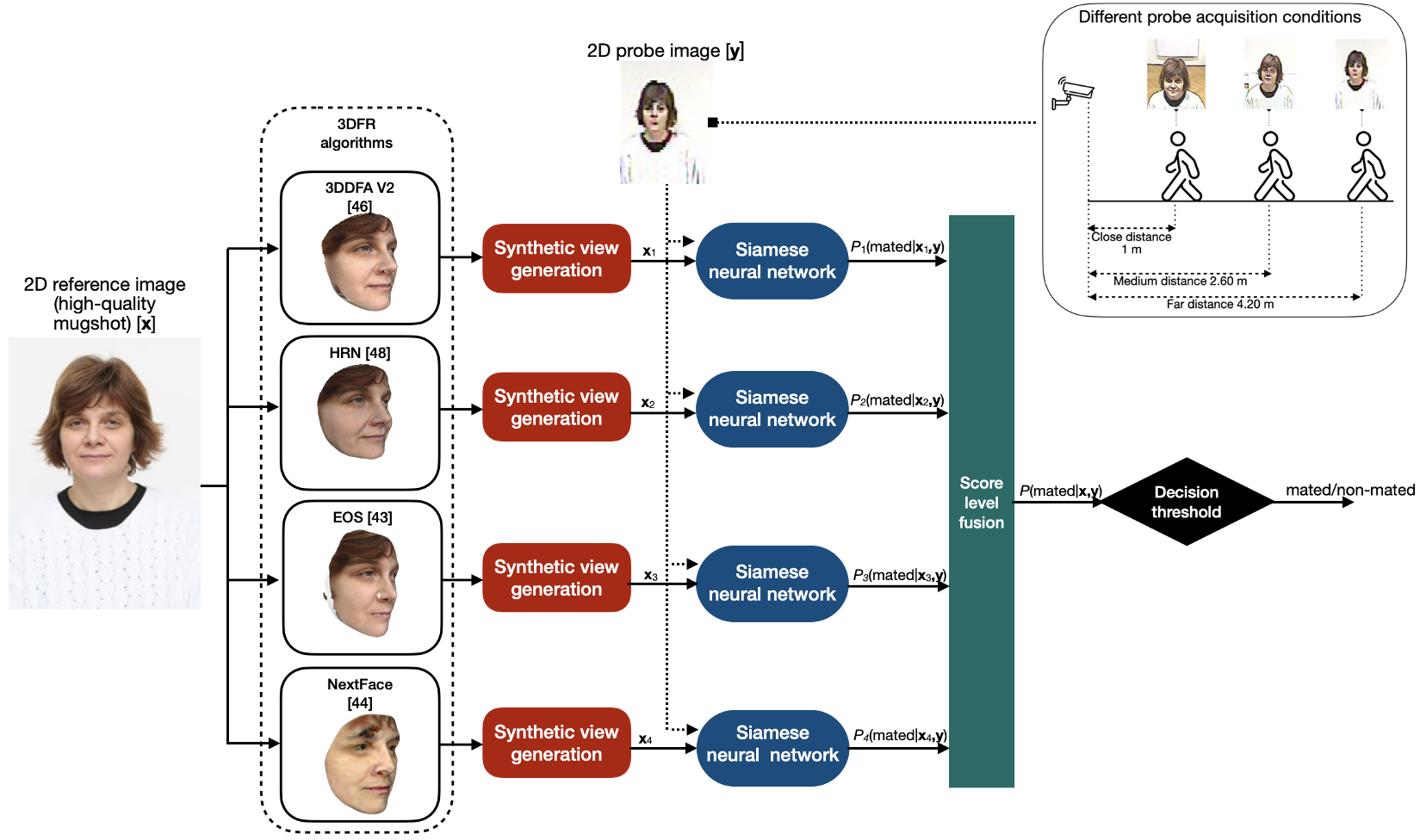}
\caption{Proposed method. After the 3DFR algorithms, the synthetic view generation module creates 2D images ($\textbf{x}_i$) from the 3D templates derived from the 2D reference image. These synthetic views represent different perspectives of the reference face to enhance recognition. The 2D probe image ($\textbf{y}$) is matched against each representation to determine similarity. In addition to our proposed 3DFR fusion scheme, we also depict the main elements of our experimental setup based on the SCface dataset together with example high-quality reference and low-quality probe images at the three distances considered (probe images after face detection). For more details of the experimental setup see the original dataset description \cite{grgic2011scface} and related works \cite{tome2013,tome2015}.
}\label{fig:proposedsystem}
\end{figure*}

Figure \ref{fig:proposedsystem} provides a graphical representation of our proposed method to investigate the potential of combining multiple 3DFR algorithms in uncontrolled scenarios. The system is mainly proposed to enhance face verification, thus facilitating the task of determining whether the identity in a probe image matches the one represented in the reference data (\textit{i.e.}, mugshot or template). Next, we describe each of the represented modules.

%uncomment

\subsection{3DFR Algorithms}\label{subsec:algorithms}
\begin{figure*}[]
	\centering
	\includegraphics[width=0.75\linewidth]{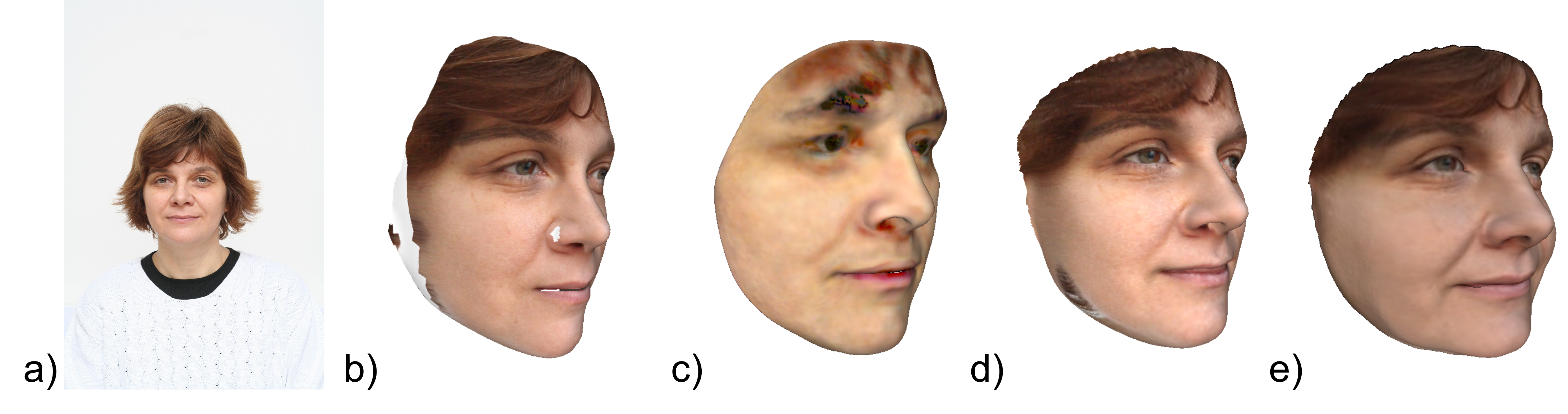}
	\caption{Examples of personalized 3D templates generated from a mugshot (a) in the SCface dataset \cite{grgic2011scface} using EOS \cite{eos} (b), NextFace \cite{dib2021practical} (c), 3DDFA V2 \cite{guo2020towards}, and HRN \cite{lei2023hierarchical} (e).}
	\label{fig:3Dmodels}
\end{figure*}

We select four state-of-the-art 3DFR algorithms, each representative of one or a combination of the previously described reconstruction approaches in order to build the 3D templates from high-quality reference data (\textit{i.e.}, frontal mugshot images).
\begin{itemize}
    \item \textit{EOS} \cite{eos}: This algorithm is based on a statistical method based on 3DMM, mainly developed for reconstructing 3D faces in time-critical applications (Figure \ref{fig:3Dmodels}b)\footnote{\url{https://github.com/patrikhuber/eos}}.
    \item \textit{NextFace} \cite{dib2021practical}: This algorithm combines a statistical approach with a photometric one for making 3D reconstruction robust to lighting conditions (Figure \ref{fig:3Dmodels}c)\footnote{\url{https://github.com/abdallahdib/NextFace}}. Specifically, the 3D face is modeled through a coarse-to-fine approach, relying on a 3DMM for a coarse extraction of the geometry and albedo, then refining the model based on the photo consistency loss between the ray-traced image and the real one and minimizing the skin reflectance through the Cook-Torrance bidirectional scattering distribution function \cite{schlick1994inexpensive}.
    \item \textit{3DDFA V2} \cite{guo2020towards}: This algorithm is based on a lightweight deep learning network for regressing the 3DMM parameters and is mainly used for 3D dense face alignment (Figure \ref{fig:3Dmodels}d)\footnote{\url{https://github.com/cleardusk/3DDFA_V2}}. In particular, it is based on MobileNet \cite{howard2017mobilenets}, a Convolutional Neural Network (CNN) optimized for mobile vision applications.
    \item \textit{HRN} \cite{lei2023hierarchical}: The HRN algorithm combines photometric features, statistical model fitting, and deep learning to reconstruct 3D facial models from in-the-wild 2D images (Figure \ref{fig:3Dmodels}e)\footnote{\url{https://github.com/youngLBW/HRN}}. In particular, it employs a coarse-to-fine approach based on two pix2pix networks \cite{isola2017image} for reconstructing a rough 3D model through 3DMM and photometric features, then refining it through a deformation map in the UV space and a displacement map to include local mid-frequency and high-frequency details. 
 
\end{itemize}

\subsection{3DFR-based Enhancement} We focus on view-based approaches to enhance face verification through 3DFR. Specifically, we train multiple face recognition systems using a gallery enlargement strategy involving a distinct 3DFR algorithm for each neural network to improve robustness against pose variations. This involves projecting each 3D template generated from training mugshots into multiple view angles, following the algorithm proposed in \cite{la2024exploring} and reported in the \textit{Supplemental Material}. Each neural network is trained with these diverse representations to extract relevant information for face verification from non-frontal poses. For testing, we generate only a frontal face for each subject in the test set, which is then compared with the corresponding set of probe images.
Notably, the computational cost introduced by the gallery enlargement strategy is primarily incurred offline, making it a minor issue for many of the intended application scenarios. This allows the system to remain efficient and practical in real-time operations \cite{la20233d}. Despite being out of scope for this proposal, future studies could explore other possible trade-offs between performance enhancement and computational complexity, such as integrating a face detection and adaptation step or comparing against multiple views of the 3D templates.

\subsection{Face Recognition Methods} 
We consider four popular deep learning networks with different computational complexities to simulate various application scenarios with distinct constraints, aiming to vary trade-offs between performance, available computational resources, and response time. In particular, we propose to use Siamese networks, which have demonstrated effective face recognition from low-resolution images \cite{lai2021deep}. 
Accordingly, we perform different experiments, employing one between two lightweight networks, namely MobileNet V3 Small \cite{howard2019searching} and XceptionNet \cite{chollet2017xception}, and a more complex one, namely VGG19 \cite{simonyan2014very}, as the backbone of each Siamese network. To further validate the contributions of our approach, we include AdaFace \cite{kim2022adaface}, a state-of-the-art face recognition model specifically designed to deal with challenging scenarios such as extreme poses, occlusions, or low-resolution images. The latter model is based on the InsightFace-ResNet101 (IR101) \cite{deng2019arcface} backbone model, a Residual Network tailored for face recognition tasks, balancing efficiency and high accuracy with moderate parameter complexity.
In summary, we explore four different Siamese networks based on the models mentioned before.

For the first three backbones (MobileNet V3 Small, XceptionNet, and VGG19), we compute the \textit{a posteriori} probability \( P(\text{mated}|\textbf{x, y}) \) to determine if the identities of the reference image \( \textbf{x} \) and the probe image \( \textbf{y} \) match. This is achieved by evaluating the similarity between their feature embeddings \( \textbf{f}(\textbf{x}) \) and \( \textbf{f}(\textbf{y}) \) using the Euclidean distance \( d_\text{E} \) as follows:

\begin{footnotesize}
\begin{center}
\begin{equation}
  P(\text{mated}|\textbf{x},y) = \frac{1}{d_\text{E}(\textbf{f}(\textbf{x}),\textbf{f}(\textbf{y}))+1}
\end{equation}
\end{center}
\end{footnotesize}

In particular, we estimate the \textit{a posteriori} probability through the previous formula to limit the range to $[0, 1]$. Concerning the AdaFace system, we calculate $P(\text{mated}|\textbf{x},\textbf{y})$ by means of the Cosine similarity\footnote{\url{https://github.com/leondgarse/Keras_insightface}} between the pair of embeddings $\textbf{f}(\textbf{x})$ and $\textbf{f}(\textbf{y})$:

\begin{footnotesize}
\begin{equation}
P(\text{mated}|\textbf{x},\textbf{y}) = \frac{\textbf{f}(\textbf{x}) \cdot \textbf{f}(\textbf{y})}{\|\textbf{f}(\textbf{x})\| \|\textbf{f}(\textbf{y})\|}
\end{equation}
\end{footnotesize}

\subsection{Fusion Methods}\label{subsec:fusions}
\begin{figure}[t]
\centering
\includegraphics[width=0.6\linewidth]{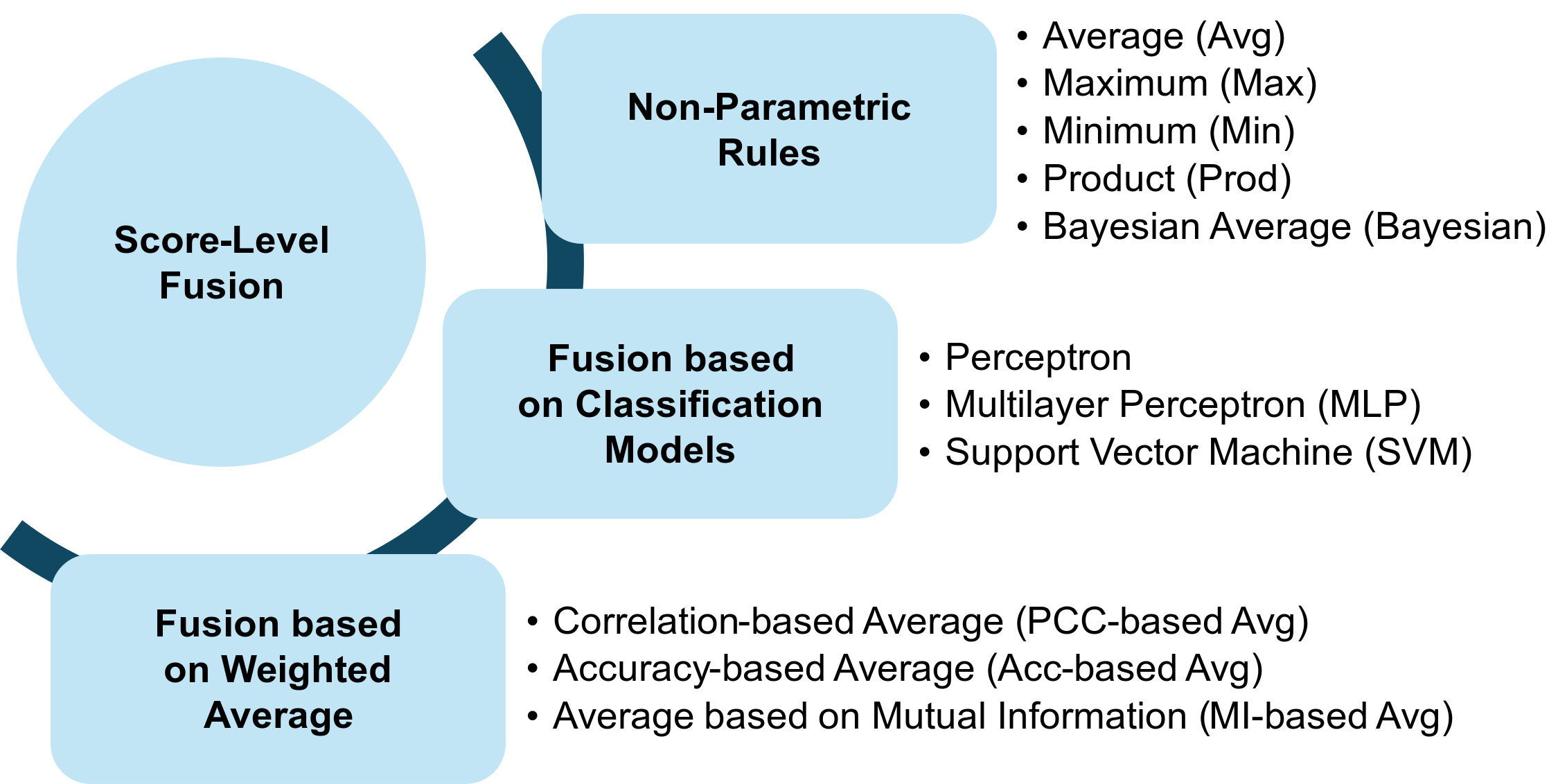}
\caption{Taxonomy of the analyzed score-level fusion methods.}\label{fig:fusionrules}
\end{figure}

Despite significant advancements in 3DFR and face recognition, there is a notable gap in the literature regarding score-level fusion between face recognition systems enhanced through various 3DFR algorithms. Accordingly, we aim to systematically evaluate which score-level fusion approaches best suit this task. Specifically, we investigate fusion methods categorized into parametric and non-parametric methods. The proposed parametric methods are further divided into weighted combination and classification models \cite{concas2022analysis}. Figure \ref{fig:fusionrules} provides a summary of all the analyzed score-level fusion methods.

In the following sections, we present a selected and representative subset of fusion methods. Additional definitions and results for other methods are provided in the \textit{Supplemental Material} to maintain the conciseness of the primary discussion. 

\subsubsection{Non-Parametric Fusion Rules}

Let us consider the fusion of $N$ classifier scores, where $P_i(\text{mated}|\textbf{x},\textbf{y})$ represents the score of the $i$-th classifier on the comparison pair $\textbf{x}$ and $\textbf{y}$. According to this notation, it is possible to compute the fusion scores using the following formulas:

\begin{footnotesize}
\begin{center}
\begin{equation}
P_\text{avg} = \frac{1}{N}\sum\limits_{i = 1}^N {P_i(\text{mated}|\textbf{x},\textbf{y})} \label{eq}
\end{equation}
\end{center}
\begin{center}
\begin{equation}
P_\text{bayes} = \frac{{\prod_{i=1}^{N}{P_i(\text{mated}|\textbf{x},\textbf{y}))}}}{{\prod_{i=1}^{N}{P_i(\text{mated}|\textbf{x},\textbf{y})) + \prod_{i=1}^{N}{(1 - P_i(\text{mated}|\textbf{x},\textbf{y}))}}}}  
\end{equation}
\end{center}
\end{footnotesize}

They represent, respectively, the simple average between the scores obtained from the single recognition systems and the Bayesian average between them.

\subsubsection{Fusion Methods based on Weighted Combinations}
Considering the advantages of weighted score-level combinations in other research on facial biometrics \cite{concas2022analysis}, we also evaluate here fusion methods based on weighted averages. The baselines of face recognition and 3DFR models are often known, and some methods are more efficient than others in specific application contexts. Assigning them a higher weight and, therefore, greater relative importance in the fusion process could allow the complementarity to be exploited without losing precision on correctly classified samples.
In general, fusion methods based on the weighted average use weights assigned to each matcher $w_i$ to combine the scores assigned to each sample linearly:

\begin{footnotesize}
\begin{center}
\begin{equation}
P_\text{weight} = \frac{\sum\limits_{i = 1}^N {w_i \cdot P_i(\text{mated}|\textbf{x},\textbf{y})}}{\sum\limits_{i = 1}^N {w_i}} \label{eq}
\end{equation}
\end{center}
\end{footnotesize}

There are various ways of assigning such weights. Here, we introduce a method based on correlation while we include additional methods reported in Figure \ref{fig:fusionrules} in the \textit{Supplemental Material}. Specifically, through this correlation-based method, named  \textit{PCC-based Avg}, we assign the weights to the fusion module as the Pearson Correlation Coefficients between the scores $P_i(\text{mated}|\textbf{x},\textbf{y})$ predicted by the single 3DFR-enhanced face verification systems from the validation samples and the actual classes ($\text{mated}$ or $\text{non-mated}$). Hence, the higher the correlation between a matcher's scores and the target, the greater its weight in the fusion.

\subsubsection{Fusion Methods based on Classification Models}

As already proposed in previous studies in the biometric field \cite{Ross2009,concas2022analysis}, trained models could effectively fuse the matching scores generated by multiple face recognition systems to determine the class label. 
In this case, we jointly consider all the outputs of the \textit{N} systems being fused $[P_1,\ldots,P_N]$ as a vector input to a classifier to determine the final probability that the pairs of reference and probe images $\textbf{x}$ and $\textbf{y}$ present the same identity.

In particular, we consider a logit-based perceptron as a stacked fusion rule, trained by minimizing a cross-entropy cost function over the scores obtained from the validation samples employed for the specific 3DFR-enhanced face recognition system. Further classification models are presented in the \textit{Supplemental Material}.

\section{Experimental Framework}\label{sec:framework}

The experimental framework is outlined in three parts. Section \ref{subsec:database} reports the datasets and settings used for the experiments. Section \ref{subsec:setup} explains the experimental protocol regarding the setup considered for the face recognition systems. It is important to note that, as mentioned in Section \ref{sec:proposed}, no additional data is used to retrain the parameters of the 3DFR modules (\textit{i.e.}, we consider the original versions of the 3DFR algorithms from the corresponding GitHub repositories). Finally, Section \ref{subsec:performance} describes the performance metrics assessed in the analyzed settings.

\subsection{Datasets and Experimental Protocol}\label{subsec:database}

State-of-the-art face recognition systems, including those based on 2D methods, achieve near-perfect performance on traditional benchmark databases \cite{fuad2021recent,melzi2024frcsyn}. However, the key advantage of 3DFR algorithms emerges in more challenging scenarios, such as those encountered in surveillance applications \cite{la20223d,proencca2018trends,sosa2018}.
Therefore, to validate our proposal, we employed the SCface dataset \cite{grgic2011scface,tome2013,tome2015combination}. This dataset contains high-quality mugshot-like images and lower-quality RGB and grayscale surveillance images of 130 subjects. The surveillance images were acquired through different camera models at three varying distances from the subject, ranging from 1 to 4.2 meters. Each
individual was captured once for every possible camera-distance combination, with the camera slightly above the subject's head, as is typical of surveillance footage \cite{castro2021systematic}.
SCface is ideal for analyzing how different camera qualities and resolutions affect face recognition performance and the system's robustness at varying distances \cite{la20233d,sosa2018person}. This makes the dataset suitable for ``cross-settings'' experiments, which test recognition across camera types or distances. Conversely, experiments using the same camera and distance for training and testing are referred to as ``intra-settings'' experiments.

We use RGB samples from 25 subjects (approximately 20\% of the dataset) as the test set for both protocols. The remaining 105 subjects are divided into training and validation sets, with 90\% used for training and 10\% for validation. We randomly divide the data for the training and test sets to minimize potential bias in experiments involving single camera-distance settings \cite{bias22}. Additionally, we implement a face detection step, following the procedure outlined for the SCface dataset in \cite{yang2017discriminative}.

To further validate our proposed approach, we also perform a cross-dataset analysis, testing the verification systems based on AdaFace \cite{kim2022adaface} on the Quis-Campi dataset \cite{neves2018quis,proencca2018trends}. This dataset contains reference images acquired indoors from 320 subjects and surveillance videos acquired outdoors in unconstrained conditions at about 50 meters. In addition to the distance and adverse factors like expression, occlusion, illumination, pose, motion blur, and out-of-focus, this dataset lacks a good set of reference images, making the recognition task more challenging \cite{la20233d}.

\subsection{Face Verification Systems: Setup}\label{subsec:setup}

\begin{figure}
    \centering
    \includegraphics[width=0.7\linewidth]{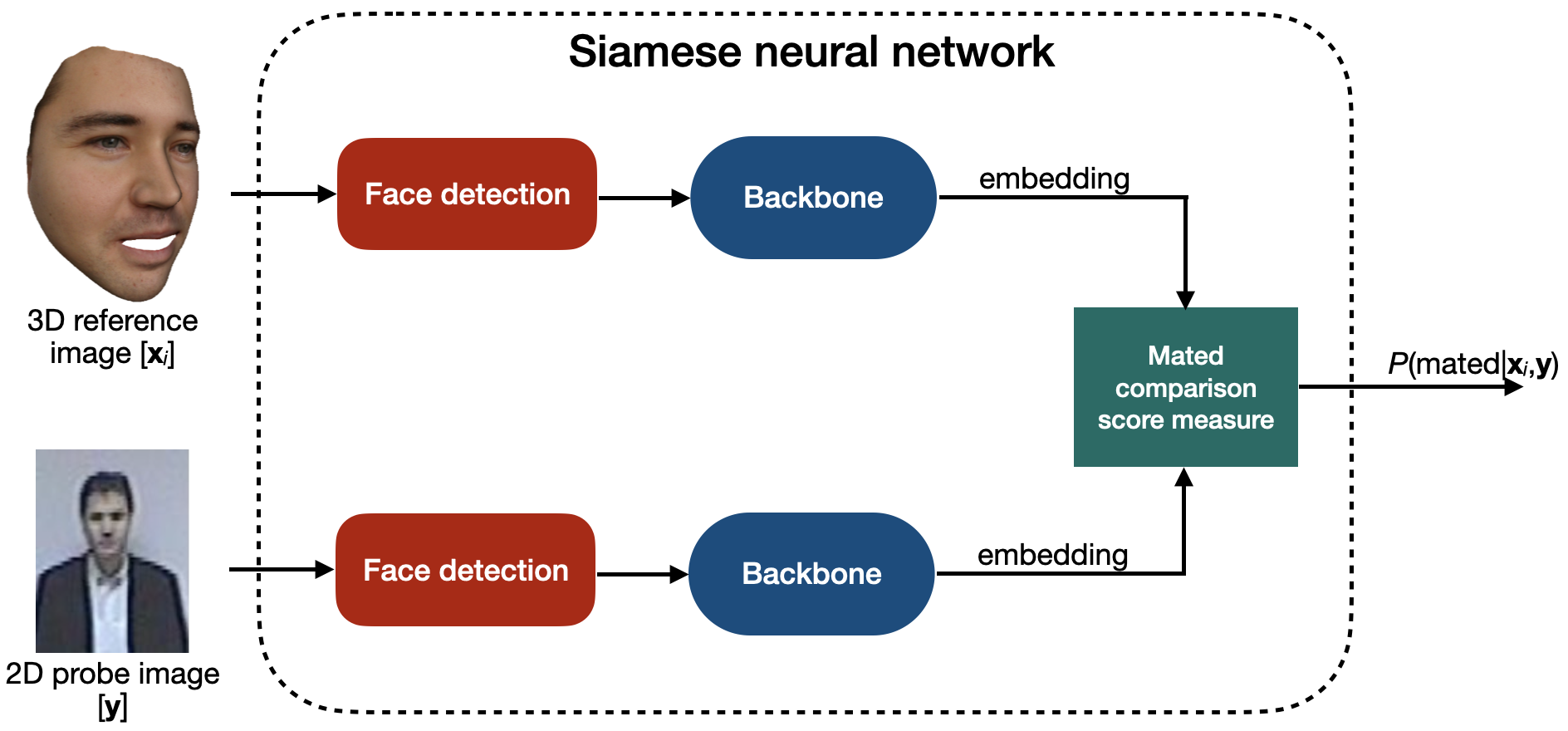}
    \caption{Siamese Neural Network module overview: the 3D reference image and 2D probe image undergo face detection followed by feature extraction through a shared backbone network. Note that, during testing, we only generate frontal faces from the reference image, which are then compared with the corresponding set of probe images. The embeddings are compared using a mated comparison score measure to compute the likelihood $P(\text{mated}|\textbf{x}_i,\textbf{y})$, indicating whether the two inputs correspond to the same individual. }
    \label{fig:subsystem_verification}
\end{figure}

All Siamese networks are pre-trained on the LFW dataset \cite{huang2008labeled}, followed by fine-tuning using the SCFace training set, with early stopping based on the validation performance on the validation set, as described in Section \ref{subsec:database}. As illustrated in Figure \ref{fig:subsystem_verification}, the networks take as input a probe image and the synthetic view generated from a 3D template corresponding to the claimed identity. For consistency, all models are trained using batches of $128$ comparisons of images, resampled at a resolution of $128 \times 128$. Training is conducted for up to $256$ epochs, with early stopping triggered after $5$ epochs of no improvement. 
All the experiments involve a learning rate of $0.001$. The models based on AdaFace are trained using the SGDW optimizer \cite{kim2022adaface}, while the other models are trained using the Adam optimizers \cite{la2024exploring}.

It is important to note that this setup is not necessarily optimized for the best performance. It is a general configuration that can be refined and improved in future research.

\subsection{Performance Evaluation and Metrics} \label{subsec:performance}

After training the individual face recognition systems using different 3DFR algorithms, we first conduct a correlation analysis to explore the potential benefits of combining these systems. Specifically, we analyze the linear correlation between pairs of face recognition systems with the same network architecture but enhanced by different 3DFR algorithms. The aim is to assess the complementary information provided by each 3DFR algorithm, using the Pearson correlation coefficient (PCC) as a key metric.

Next, we conduct two sets of experiments to evaluate the effectiveness of fusion methods - combining face recognition systems trained on different 3DFR algorithms - in the RGB domain. The experiments are divided into two protocols: intra-setting and cross-setting. The intra-setting protocol involves training and testing the systems using data captured by a specific camera at a fixed distance, allowing us to assess model performance under controlled conditions. In contrast, the cross-setting protocol involves training and testing the systems on images captured by different cameras and/or at varying distances to evaluate the generalizability of the systems across different acquisition settings. The latter allows us to individually and jointly observe the effect of the variation regarding the expected camera and the distance involved in the acquisition.

To assess the robustness of the fusion methods, we consider several metrics commonly used in face recognition. From the matching scores, we calculate the False Match Rate (FMR) - the percentage of non-matching pairs incorrectly classified as matches - and the False Non-Match Rate (FNMR) - the percentage of matching pairs misclassified as non-matches - at varying thresholds. These enable us to generate Receiver Operating Characteristic (ROC) curves by plotting FMR against $1 - \text{FNMR}$ and calculating the Area Under the Curve (AUC) to measure overall performance.

Similarly, from the distributions of the scores related to matching and non-matching identities, we assess the effect size through Cohen's $d$ to further highlight the discriminatory ability of the evaluated systems. Indeed, even if two verification systems report similar AUC values, Cohen's $d$ can also provide information about the magnitude of the differences between the score distributions. The higher the Cohen's $d$ is, the more significant the differences between such distributions are.

For completeness, we also include the Equal Error Rate (EER), a widely used metric that reflects the point at which the rates of false matches and false non-matches are equal. Finally, we include the $\%$\text{FNMR}$@$\text{FMR}=$1\%$ and $\%$\text{FMR}$@$\text{FNMR}=$1\%$ to assess performance under stringent accuracy constraints, such as high-security and high-usable applications, respectively.

\section{Results}\label{sec:results}
This section presents the experimental results. Section \ref{subsec:corr} analyzes the correlation between individual classification models, each enhanced with different 3DFR algorithms. Section \ref{subsec:intra_results} discusses the findings from the intra-setting analysis. Section \ref{subsec:cross_results} details the performance outcomes from the cross-setting analysis. Lastly, Section \ref{subsec:crossdataset_results} reports the results of the cross-dataset analysis.

\subsection{Correlation Analysis}\label{subsec:corr}

\begin{figure}[t]
\centering
\includegraphics[width=0.7\linewidth]{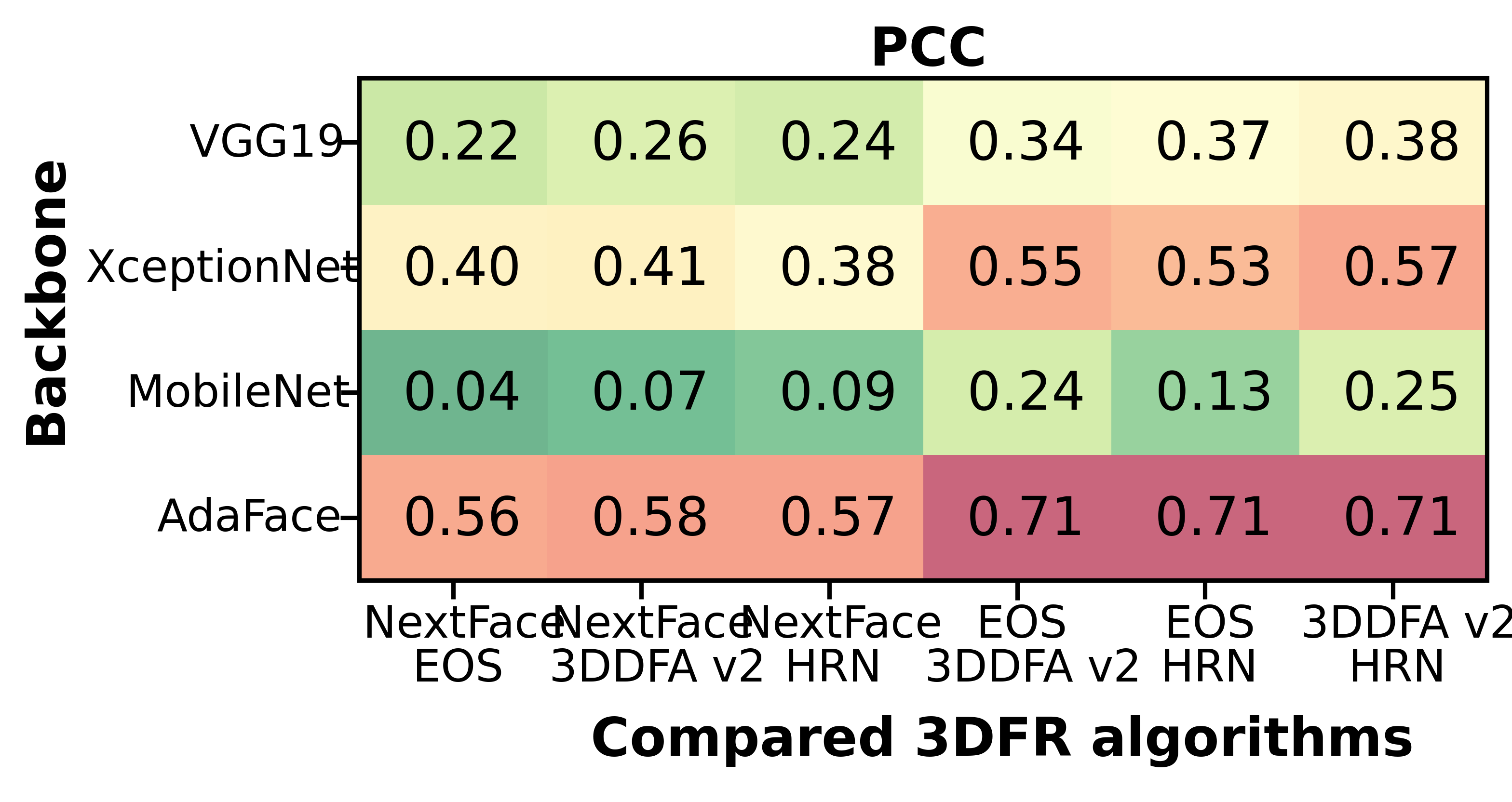}
\caption{Pearson Correlation Coefficient (PCC) between the sets of scores obtained in the intra-setting experiments from pairs of face recognition systems based on the same backbone but enhanced by different 3DFR algorithms. Each row presents the results for one architecture, while each column presents the results obtained by comparing a pair of 3DFR algorithms.}\label{fig:corr}
\end{figure}

Figure \ref{fig:corr} shows a higher linear correlation between the verification systems enhanced by the 3DFR algorithms that generate visually better 3D templates (Figure \ref{fig:3Dmodels}). This confirms the previous observations in \cite{la2024exploring}. Still, even in these cases, the correlation is mainly weak. The only strong correlation resulting from the experiments is obtained through AdaFace, with a PCC coefficient of 0.71.

This outcome enforces the hypothesis that different 3DFR algorithms could provide distinct information to verification systems. However, an analysis of the effectiveness of the combination between them must be performed since a low correlation could be related to a relevant difference in the overall performance or mostly to differences in single types of errors. In the latter case, the information provided by each 3DFR algorithm could be effectively exploited in verification scenarios by combining them.

\subsection{Intra-Setting Analysis}\label{subsec:intra_results}

Table \ref{tab:intraRGB} reports the average performance values for individual 3DFR algorithms and their fusion in the intra-setting scenario (\textit{i.e.}, fixed camera-distance settings). The best values for each metric and architecture are highlighted in bold for both the individual 3DFR-enhanced models and the fusion approaches.

\begin{table*}[t]
\centering
\caption{\textbf{Average results in the intra-setting context.} For each metric and verification system (based on VGG19, XceptionNet, MobileNet V3, and AdaFace), we highlight the best result achieved when enhanced through the single 3DFR algorithm (3DDFA v2, EOS, NextFace, and HRN) and the fusion one (Avg, Bayesian Avg, PCC-based Avg, and Perceptron).}
\resizebox{0.95\textwidth}{!}{%
\begin{tabular}{cccccc|ccccc|}
\cline{2-11}
 &
  \multicolumn{5}{|c|}{\textbf{VGG19-based \cite{simonyan2014very} }} &
  \multicolumn{5}{c|}{\textbf{XceptionNet-based \cite{chollet2017xception}}}  \\ \hline
\multicolumn{1}{|c|}{\multirow{2}{*}{\textbf{Method}}} &
  \textbf{AUC} &
  \textbf{EER} &
  \textbf{Cohen's} &
  \textbf{\%FMR at} &
  \textbf{\%FNMR at} &
  \textbf{AUC} &
  \textbf{EER} &
  \textbf{Cohen's} &
  \textbf{\%FMR at} &
  \textbf{\%FNMR at}  \\
\multicolumn{1}{|c|}{} &
   \textbf{{[}\%{]}} &
   \textbf{{[}\%{]}} &
  \textbf{d} &
  \textbf{FNMR=1\%} &
  \textbf{FMR=1\%} &
   \textbf{{[}\%{]}} &
   \textbf{{[}\%{]}} &
  \textbf{d} &
  \textbf{FNMR=1\%} &
  \textbf{FMR=1\%}  \\ \hline
\multicolumn{1}{|c|}{\textbf{Baseline}} &
  75.56 &
  29.31 &
  0.94 &
  87.31 &
  87.03 &
  76.03 &
  25.86 &
  0.89 &
  83.93 &
  90.91  \\ \hline
\multicolumn{1}{|c|}{\textbf{3DDFA v2 \cite{guo2020towards}}} &
  81.56 &
  20.94 &
  1.07 &
  74.93 &
  \textbf{90.69} &
  80.35 &
  24.38 &
  1.02 &
  \textbf{79.05} &
  \textbf{89.01}  \\
\multicolumn{1}{|c|}{\textbf{EOS \cite{eos}}} &
  80.63 &
  22.11 &
  1.02 &
  81.12 &
  95.04 &
  \textbf{81.55} &
  \textbf{22.29} &
  \textbf{1.07} &
  83.07 &
  89.45  \\
\multicolumn{1}{|c|}{\textbf{NextFace \cite{dib2021practical}}} &
  74.02 &
  27.21 &
  0.80 &
  71.18 &
  95.72 &
  76.16 &
  26.79 &
  0.83 &
  86.15 &
  93.47 
   \\ 
  \multicolumn{1}{|c|}{\textbf{HRN \cite{lei2023hierarchical}}} &
  \textbf{83.21} &
  \textbf{19.97} &
  \textbf{1.17} &
  \textbf{68.86} &
  92.01 &
  80.47 &
  23.83 &
  1.02 &
  80.86 &
  89.70  \\\hline
\multicolumn{1}{|c|}{\textbf{Fusion (Avg)}} &
  87.35 &
  18.78 &
  \textbf{1.60} &
  50.76 &
  84.75 &
  85.22 &
  19.05 &
  1.39 &
  61.15 &
  \textbf{84.34}  \\
\multicolumn{1}{|c|}{\textbf{Fusion (Bayesian Avg)}} &
  85.92 &
  21.86 &
  1.43 &
  56.23 &
  86.26 &
  85.13 &
  18.92 &
  1.38 &
  61.04 &
  84.85  \\
  \multicolumn{1}{|c|}{\textbf{Fusion (PCC-based Avg)}} &
    \textbf{87.43} &
    18.65 &
    1.59  &
    \textbf{49.51} &
    \textbf{84.59} &
    \textbf{85.39} &
    \textbf{18.58} &
    1.39 &
    \textbf{60.80} &
    85.07 \\
  \multicolumn{1}{|c|}{\textbf{Fusion (Perceptron)}} &
  85.83 &
  \textbf{17.80} &
  1.52 &
  58.03 &
  87.16 &
  83.09 &
  19.11 &
  \textbf{1.45} &
  69.85 &
  87.21  \\
 \hline
 \\ \cline{2-11}
  &
  \multicolumn{5}{|c|}{\textbf{MobileNet-based \cite{mobilenetv3}}} &
  \multicolumn{5}{c|}{\textbf{AdaFace \cite{kim2022adaface}}}  \\ \hline
\multicolumn{1}{|c|}{\multirow{2}{*}{\textbf{Method}}} &
  \textbf{AUC} &
  \textbf{EER} &
  \textbf{Cohen's} &
  \textbf{\%FMR at} &
  \textbf{\%FNMR at} &
  \textbf{AUC} &
  \textbf{EER} &
  \textbf{Cohen's} &
  \textbf{\%FMR at} &
  \textbf{\%FNMR at}  \\
\multicolumn{1}{|c|}{} &
   \textbf{{[}\%{]}} &
   \textbf{{[}\%{]}} &
  \textbf{d} &
  \textbf{FNMR=1\%} &
  \textbf{FMR=1\%} &
   \textbf{{[}\%{]}} &
   \textbf{{[}\%{]}} &
  \textbf{d} &
  \textbf{FNMR=1\%} &
  \textbf{FMR=1\%}  \\ \hline
\multicolumn{1}{|c|}{\textbf{Baseline}} &
  61.50 &
  50.00 &
  0.38 &
  93.65 &
  94.03 &
  93.09 &
  5.30 &
  2.99 &
  28.32 &
  30.21  \\ \hline
\multicolumn{1}{|c|}{\textbf{3DDFA v2 \cite{guo2020towards}}} &
  51.10 &
  \textbf{50.00} &
  0.10 &
  96.88 &
  98.83 &
  97.98 &
  2.89 &
  3.49 &
  \textbf{14.88} &
  20.06  \\
\multicolumn{1}{|c|}{\textbf{EOS \cite{eos}}} &
  49.97 &
  \textbf{50.00} &
  0.10 &
  97.23 &
  \textbf{98.37} &
  \textbf{98.25} &
  \textbf{1.91} &
  \textbf{3.67} &
  17.44 &
  \textbf{17.01}  \\
\multicolumn{1}{|c|}{\textbf{NextFace \cite{dib2021practical}}} &
  \textbf{52.48} &
  \textbf{50.00} &
  0.10 &
  97.70 &
  98.58 &
  96.04 &
  8.31 &
  2.69 &
  33.33 &
  35.02 
   \\ 
  \multicolumn{1}{|c|}{\textbf{HRN \cite{lei2023hierarchical}}} &
  51.94 &
  \textbf{50.00} &
  \textbf{0.13} &
  \textbf{96.47} &
  98.81 &
  97.99 &
  3.39 &
  3.54 &
  17.62 &
  19.83  \\\hline
\multicolumn{1}{|c|}{\textbf{Fusion (Avg)}} &
  51.25 &
  \textbf{50.00} &
  0.09 &
  \textbf{96.72} &
  99.28 &
  98.99 &
  \textbf{0.68} &
  4.18 &
  10.07 &
  \textbf{10.54}  \\
\multicolumn{1}{|c|}{\textbf{Fusion (Bayesian Avg)}} &
  50.00 &
  \textbf{50.00} &
  0.00 &
  100.00 &
  100.00 &
  98.99 &
  \textbf{0.68} &
  4.56 &
  10.16 &
  10.96  \\
  \multicolumn{1}{|c|}{\textbf{Fusion (PCC-based Avg)}} &
    \textbf{51.52} &
    \textbf{50.00} &
    0.07  &
    97.75 &
    \textbf{98.33} &
    \textbf{99.00} &
    \textbf{0.68} &
    4.20 &
    \textbf{9.92} &
    10.77 \\
  \multicolumn{1}{|c|}{\textbf{Fusion (Perceptron)}} &
  50.17 &
  \textbf{50.00} &
  \textbf{0.15} &
  97.52 &
  99.77 &
  98.95 &
  0.80 &
  \textbf{8.93} &
  10.01 &
  10.78  \\
 \hline
\end{tabular}%
}
\label{tab:intraRGB}
\end{table*}

\begin{figure*}[t]
\centering
\includegraphics[width=\linewidth]{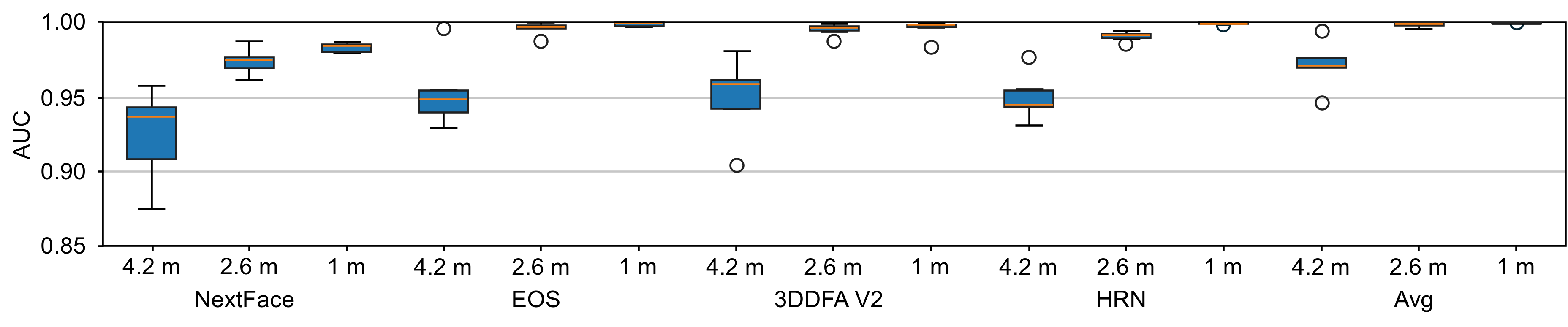}
\caption{Summary of AUC values obtained from all the intra-setting camera-distance configurations (\textit{i.e.}, 4.2 meters, 2.6 meters, and 1 meter) through AdaFace trained on single 3DFR algorithms, and fusion of them through the Average rule (Avg).
}\label{fig:intra_dist_boxplot}
\end{figure*}

First of all, the results confirm that 3DFR algorithms are able to improve the performance of a deep learning-based verification system in surveillance scenarios. In particular, this result is confirmed by using an architecture specifically proposed for recognition in acquisition conditions representative of such scenarios (\textit{i.e.}, AdaFace). However, it is important to note that the global improvement (AUC and Cohen's $d$) is not necessarily confirmed at all thresholds, leading, for example, to a worse \%FNMR at FMR=1\% in the VGG19-based architecture compared to the model trained only on mugshots (\textit{i.e.}, 87.03\% vs. 90.69\%). Moreover, it is important to note that recognition capability can only improve when using a system with performance that is not close to that of random selection, such as the MobileNet-based architecture. For this reason, we will focus on the other three architectures in the following analyses.

At the single architecture level, we can observe that although there is a 3DFR algorithm that is most successful in improving the global average performance, this improvement does not appear to be consistent for all decision thresholds (\textit{e.g.}, \%FMR at FNMR=1\%), showing better performances through other 3DFR algorithms. This outcome further enforces the hypothesis of a potential complementarity between the systems enhanced through 3DFR algorithms.

Moreover, analyzing the robustness of the proposed systems with an increasing acquisition distance (\textit{i.e.}, 1 meter, 2.6 meters, and 4.2 meters, see Figure \ref{fig:intra_dist_boxplot}), we can observe that the single 3DFR algorithms show differences in enhancement capability, revealing that there is not a single best choice for all settings. For instance, on average, AdaFace enhanced by 3DDFA V2 revealed the highest performance on probe images acquired at 4.2 meters from the subjects,
while the same verification system enhanced by HRN was revealed to be the best-performing one at a 1-meter distance.

Score-level fusion results confirmed that most of the fusion methods are able to exploit such differences, revealing improved performance both at a global (\textit{i.e.}, AUC and Cohen's $d$) and local level (\textit{i.e.}, EER, \%FMR at FNMR=1\%, and \%FNNR at FMR=1\%). The only worse result is the EER obtained through the Bayesian average rule compared to the best 3DFR-enhanced VGG19-based verification system (\textit{i.e.}, 21.86\% vs. 19.97\%), which still performs worse with respect to all other performance metrics. 

Notably, fusion rules can also solve the problem introduced by integrating 3DFR algorithms in systems training on stringent decision thresholds, showing better performances than baseline systems. This result at the most stringent decision thresholds is particularly evident with AdaFace, with a notable reduction in error compared to systems based on single reconstruction models, which becomes even more significant compared to the values reported by the baseline system.

Unlike previous studies on facial biometrics \cite{concas2022analysis}, parametric fusion methods do not lead to a clear improvement over the use of non-parametric fusion rules in intra-setting scenarios. For instance, the AUC values are between 98.95\% and 99.00\% for all the reported fusions with AdaFace. From a technologist's point of view, this outcome is positive since it eliminates the requirement for setting further parameters or training the score-level fusion module any time a new system is integrated into the verification system. However, this observation is valid whenever the newly included model does not perform significantly poorly. In that case, the impact on the final system's performance would not be modulated by assigning a lower weight to the final decision by the fusion module.

\begin{figure*}[t]
\centering
\includegraphics[width=\linewidth]{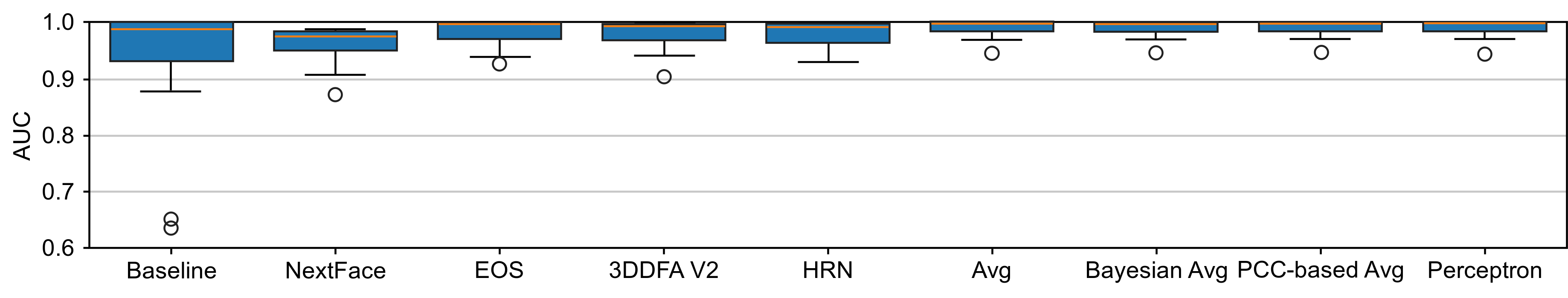}
\caption{AUC values obtained with AdaFace in the intra-setting scenario}.\label{fig:intra_boxplot}\\
\vspace{0.2cm}
\includegraphics[width=\linewidth]{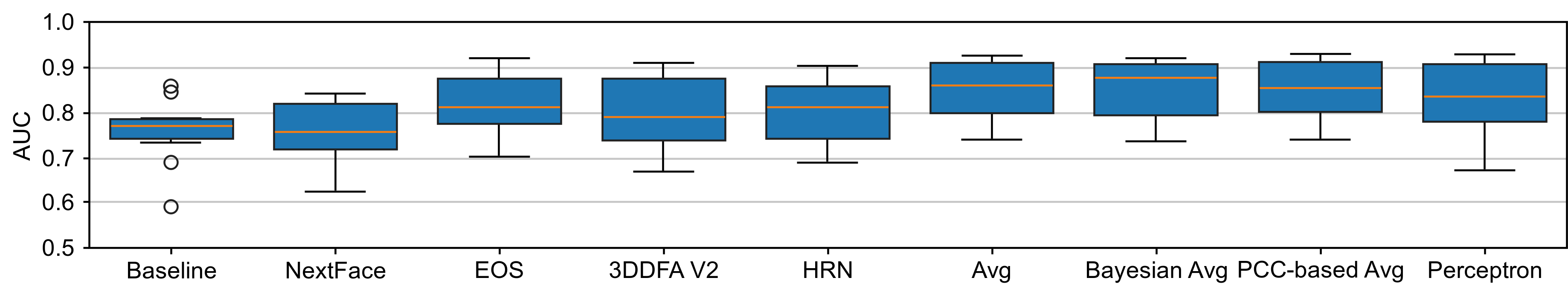}
\caption{AUC values obtained with the Siamese network based on the XceptionNet in the intra-setting scenario.}\label{fig:intra_boxplot_xcp}
\end{figure*}

Furthermore, as can be seen in Figures \ref{fig:intra_boxplot} and \ref{fig:intra_boxplot_xcp}, it is worth noting that the investigated non-parametric fusion rules and the weighted average can also deal with the variability in performance observed by the 3DFR-enhanced recognition systems between the different camera-distance settings, resulting in more stable results across the experiments.

\subsection{Cross-Setting Analysis}\label{subsec:cross_results}

Considering the unsuitability of the architecture based on MobileNet and the results previously obtained in \cite{la2024exploring}, the following discussion in the cross-settings scenario focuses on the Siamese network based on XceptionNet and AdaFace.

Table \ref{tab:CROSSRGB} reports the results of varying acquisition distances and surveillance cameras between training and test sets. As expected by previous work \cite{la2024exploring}, such a challenging cross-settings scenario leads to an overall decay in performance. However, this degradation is not consistent with all the analyzed performance metrics concerning AdaFace. Specifically, while the EER in the cross-setting experiments of the baseline model is undesirably higher than in the intra-setting scenario, such a value is lower both when the system is enhanced by a single 3DFR algorithm and after the fusion (\textit{e.g.}, from 8.42\% in the Baseline to 0.06\% with the Avg fusion).

\begin{table*}[t]
\centering
\caption{\textbf{Average results in the cross-setting context} involving both different acquisition distances and cameras between training and test set. For each metric and verification system (based on  XceptionNet and AdaFace), we highlight the best result achieved when enhanced through the single 3DFR algorithm (3DDFA v2, EOS, NextFace, and HRN) and the fusion one (Avg, Bayesian Avg, PCC-based Avg, and Perceptron).}
\resizebox{0.95\textwidth}{!}{%
\begin{tabular}{cccccc|ccccc|}
\cline{2-11}
 &
  \multicolumn{5}{|c|}{\textbf{XceptionNet-based \cite{chollet2017xception}}} &
  \multicolumn{5}{c|}{\textbf{AdaFace \cite{kim2022adaface}}  }\\ \hline
\multicolumn{1}{|c|}{\multirow{2}{*}{\textbf{Method}}} &
  \textbf{AUC}&
  \textbf{EER} &
  \textbf{Cohen's} &
  \textbf{\%FMR at} &
  \textbf{\%FNMR at} &
  \textbf{AUC} &
  \textbf{EER} &
  \textbf{Cohen's} &
  \textbf{\%FMR at} &
  \textbf{\%FNMR at}  \\
\multicolumn{1}{|c|}{} &
   \textbf{{[}\%{]}} &
   \textbf{{[}\%{]}} &
  \textbf{d} &
  \textbf{FNMR=1\%} &
  \textbf{FMR=1\%} &
   \textbf{{[}\%{]}} &
   \textbf{{[}\%{]}} &
  \textbf{d} &
  \textbf{FNMR=1\%} &
  \textbf{FMR=1\%}  \\ \hline
\multicolumn{1}{|c|}{\textbf{Baseline}} &
  68.97 &
  36.21 &
  0.63 &
  91.81 &
  96.53  &
  96.95 &
  8.42 &
  2.89 &
  47.40 &
  34.51  \\ \hline
\multicolumn{1}{|c|}{\textbf{3DDFA v2 \cite{guo2020towards}}} &
  70.95 &
  19.70 &
  \textbf{0.69} &
  96.22 &
  92.53 &
  97.05 &
  0.53 &
  3.19 &
  26.91 &
  26.82  \\
\multicolumn{1}{|c|}{\textbf{EOS \cite{eos}}} &
  \textbf{71.21} &
  \textbf{19.09} &
  0.61 &
  96.57 &
  \textbf{91.58}  &
  \textbf{97.26} &
  \textbf{0.12} &
  \textbf{3.34} &
  \textbf{23.63} &
  \textbf{22.25}  \\
\multicolumn{1}{|c|}{\textbf{NextFace \cite{dib2021practical}}} &
  68.47 &
  23.46 &
  0.57 &
  95.99 &
  94.97 &
  94.15 &
  3.57 &
  2.35 &
  40.67 &
  48.40 
   \\ 
  \multicolumn{1}{|c|}{\textbf{HRN \cite{lei2023hierarchical}}} &
  66.94 &
  22.25 &
  0.48 &
  \textbf{94.82} &
  95.63 &
  97.12 &
  1.05 &
  3.24 &
  25.44 &
  25.77  \\\hline
\multicolumn{1}{|c|}{\textbf{Fusion (Avg)}} &
  75.77 &
  \textbf{16.44} &
  0.86 &
  83.41 &
  89.84 &
  \textbf{98.55} &
  \textbf{0.06} &
  3.80 &
  13.78 &
  \textbf{16.34}  \\
\multicolumn{1}{|c|}{\textbf{Fusion (Bayesian Avg)}} &
  \textbf{77.03} &
  26.65 &
  0.90 &
  \textbf{80.96} &
  \textbf{89.72} &
  98.54 &
  \textbf{0.06} &
  4.00 &
  \textbf{13.74} &
  16.45  \\
  \multicolumn{1}{|c|}{\textbf{Fusion (PCC-based Avg)}} &
  76.95 &
  26.81 &
  0.92 &
  81.01 &
  89.94 &
  98.54 &
  \textbf{0.06} &
  4.00 &
  \textbf{13.74} &
  16.45  \\
  \multicolumn{1}{|c|}{\textbf{Fusion (Perceptron)}} &
  75.97 &
  27.36 &
  \textbf{0.95} &
  82.28 &
  90.64  &
  98.51 &
  \textbf{0.06} &
  \textbf{7.62} &
  14.06 &
  16.37  \\
 \hline

\end{tabular}%
}
\label{tab:CROSSRGB}
\end{table*}

\begin{figure*}[t]
\centering
\includegraphics[width=\linewidth]{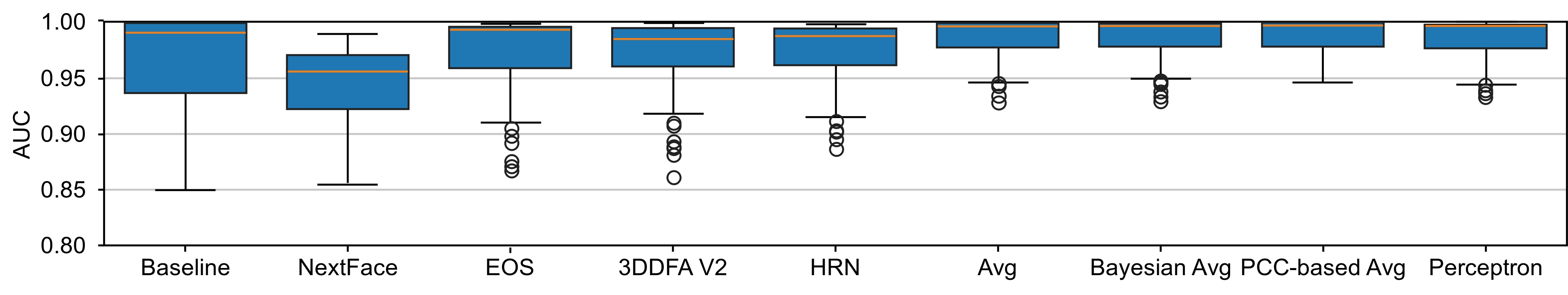}
\caption{AUC obtained with AdaFace in the cross-setting scenarios when varying acquisition distance and surveillance cameras between training and test.}\label{fig:cross_boxplot}
\end{figure*}

%Even considering the overall reduction in performance, the previous observation related to the improvements achieved from enhancing deep learning verification systems through 3DFR algorithms and the further improvement obtainable from the fusion of such systems is confirmed, as can be seen in Figure \ref{fig:cross_roc_adaface}.
Moreover, the greater stability in performance and the solution of the worst performance at certain decision thresholds after integrating 3DFR algorithms after the systems' score-level fusion are also confirmed, as well as the similarity between parametric and non-parametric score-level fusions (see Figure \ref{fig:cross_boxplot}). Analyzing the results obtained from the single cross-settings experiments involving different acquisition distances and cameras on AdaFace, one of the most relevant outcomes is that the average rule, the PCC-based average, and the fusion based on the Perception are always able to improve the AUC value compared to the best single 3DFR-enhanced verification system, confirming the contributions of the proposed approach.

%\begin{figure}[t]
%\centering
%\includegraphics[width=0.55\linewidth]{Figure/roc_adaface_crossall}
%\caption{Average ROC curves obtained with AdaFace in the cross-settings scenarios when varying both acquisition distance and surveillance cameras between training and test sets.
%}\label{fig:cross_roc_adaface}
%\end{figure}

\begin{figure}[t]
\centering
\includegraphics[width=0.6\linewidth]{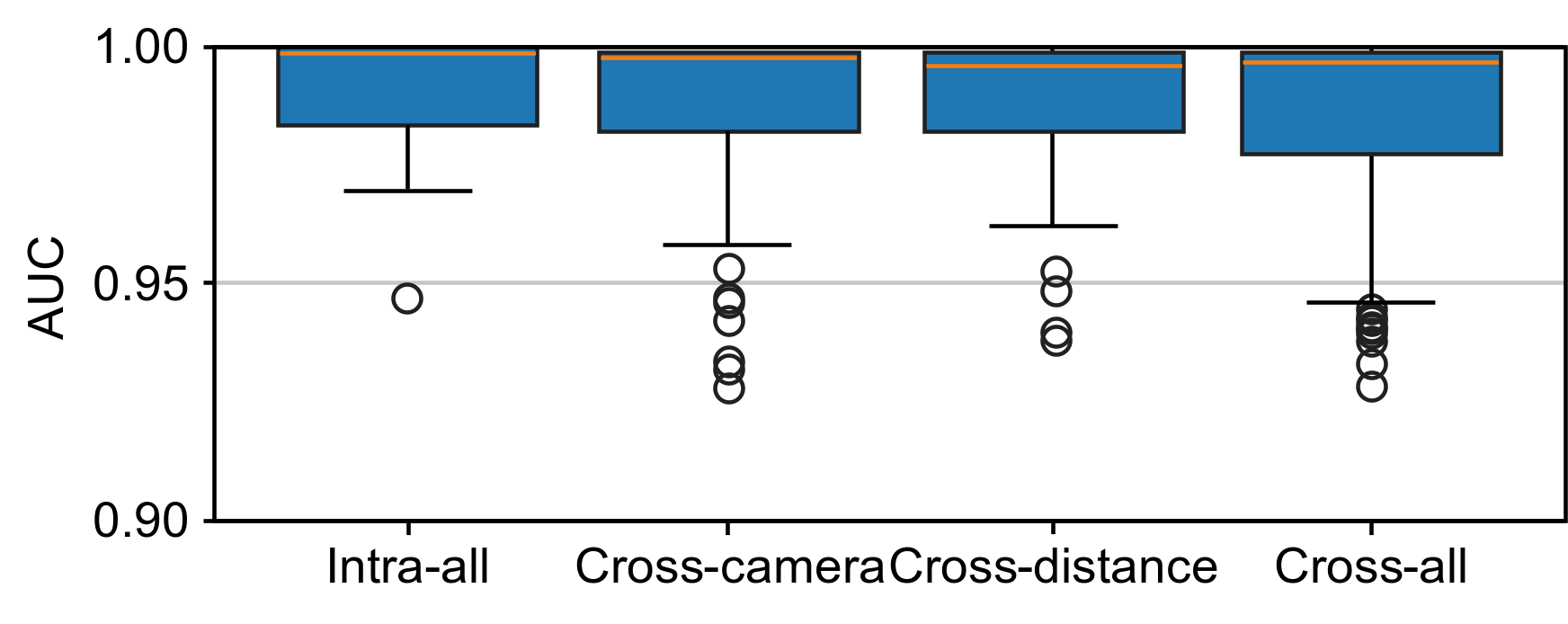}
\caption{AUC values obtained after combining the 3DFR-enhanced verification systems based on AdaFace through the average score-level fusion rule in intra-setting and cross-setting scenarios.}\label{fig:auc_comparison}
\end{figure}

Specifically, as can be seen in Figure \ref{fig:auc_comparison}, there is no clear predominance between acquisition distance and surveillance cameras regarding the adverse effect on recognition capability, causing a comparable slight decay in global performance. As expected, the effectiveness of the recognition further decreases whenever both settings vary between training and test sets. 

In addition to the pre-training with a different dataset, the robustness of the verification systems to different surveillance cameras could be related to the training of the Siamese neural networks since they deal with probe images acquired by a surveillance camera and facial representations generated from data acquired by a different camera.
This characteristic could represent an advantageous factor for the technologist, as it could justify the employment of a large realistic dataset of images for training the system, even if acquired by a surveillance camera that is different from the one employed for acquiring the probe.
Such an observation could be crucial in forensics and security, as it is not always possible for professionals to access sufficient data from the same camera model from which a probe image has been obtained \cite{la20233d}.

\begin{figure}
\centering
\subfloat[Verification on closer faces.]{\includegraphics[width=0.495\columnwidth]{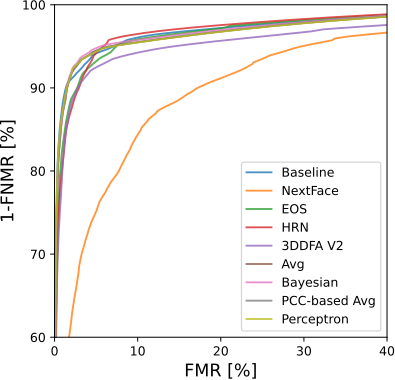}}
\hfill
\subfloat[Verification on more distant faces.]{\includegraphics[width=0.495\columnwidth]{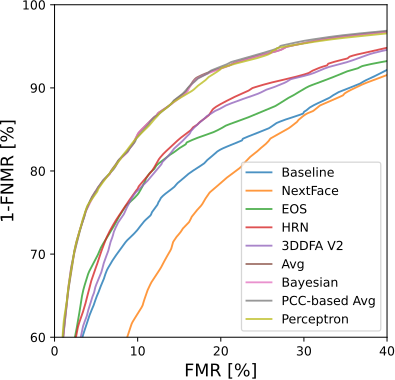}}
\caption{ROC curves obtained after combining the 3DFR-enhanced verification systems based on AdaFace obtained in the cross-setting scenario when training the system on surveillance images acquired at 4.2 meters of distance from the subjects and testing on images acquired at 1 meter of distance (a) and vice versa (b).}\label{fig:roc_cross_distance_adaface}
\end{figure}

From a deeper analysis of acquisition distance (see Figure \ref{fig:roc_cross_distance_adaface}), it is possible to observe that the reason behind the decay in performance in the cross-distance setting is mainly related to images acquired at a greater distance than the one considered in the training set. Still, this also represents the case in which score-level fusions can provide the most relevant improvement in performance, thus alleviating such an issue and contributing to the stability of the recognition performance.

%\begin{figure}[t]
%\centering
%\includegraphics[width=\linewidth]%{Figure/cross_distance.png}
%\caption{ROC curves obtained after combining the 3DFR-enhanced verification systems based on AdaFace obtained in the cross-settings scenario when training the system on surveillance images acquired at 4.2 meters of distance from the subjects and training on images acquired at 1 meter of distance (a) and vice versa (b).}\label{fig:roc_cross_distance_adaface}
%\end{figure}

\subsection{Cross-Dataset Analysis}\label{subsec:crossdataset_results}
Considering the previous outcomes in the cross-setting scenario, we also include here an analysis of the cross-dataset results obtained from the single 3DFR-enhanced verification systems based on AdaFace and their fusion through the average rule, as reported in Table \ref{tab:crossdataset_table}. 

\begin{table}[t]
\centering
\caption{Average results of the verification systems based on AdaFace in the cross-dataset context. For each metric, we highlight the best result. For the AUC, we report both average and standard deviation.}
\resizebox{0.7\columnwidth}{!}{%
\begin{tabular}{cccccc|}
\hline
\multicolumn{1}{|c|}{\multirow{2}{*}{\textbf{Method}}} &
  \textbf{AUC} &
  \textbf{EER} &
  \textbf{Cohen's} &
  \textbf{\%FMR at} &
  \textbf{\%FNMR at}  \\
\multicolumn{1}{|c|}{} &
   \textbf{{[}\%{]}} &
   \textbf{{[}\%{]}} &
  \textbf{d} &
  \textbf{FNMR=1\%} &
  \textbf{FMR=1\%}  \\ \hline
\multicolumn{1}{|c|}{\textbf{Baseline}} &
  85.60 $\pm$ 1.85 &
  19.12 &
  2.05 &
  92.31 &
  57.29 \\ \hline
\multicolumn{1}{|c|}{\textbf{3DDFA v2 \cite{guo2020towards}}} &
  86.67 $\pm$ 0.94 &
  19.29 &
  2.10 &
  88.06 &
  54.43 \\
\multicolumn{1}{|c|}{\textbf{EOS \cite{eos}}} &
  86.79 $\pm$ 0.87&
  19.44 &
  2.12 &
  97.16 &
  53.56 \\
\multicolumn{1}{|c|}{\textbf{NextFace \cite{dib2021practical}}} &
  85.78 $\pm$ 0.97 &
  21.33 &
  1.84 &
  83.02 &
  60.59 \\ 
  \multicolumn{1}{|c|}{\textbf{HRN \cite{lei2023hierarchical}}} &
  86.40 $\pm$ 0.83 &
  19.55 &
  2.10 &
  87.00 &
  54.25 \\ \hline
\multicolumn{1}{|c|}{\textbf{Fusion (Avg)}} &
  \textbf{88.72 $\pm$ 0.70} &
  \textbf{17.42} &
  \textbf{2.45} &
  \textbf{82.16} &
  \textbf{46.96} \\ \hline
\end{tabular}%
}
\label{tab:crossdataset_table}
\end{table}

As expected, such a challenging scenario involving lower-quality data concerning both probe and reference images leads to a significant decay in the overall recognition capability. Still, it is possible to observe that the enhancement of the verification systems through the 3DFR algorithms can improve the performance compared to the baseline system and that the fusion can further enhance it: 3.12\% AUC improvement between Baseline and Fusion. 

The differences in performance enhancement between distinct 3DFR algorithms align with the previous observations in Sections \ref{subsec:intra_results} and \ref{subsec:cross_results}. 
The results also confirm the further improvement in the recognition capability after the score-level fusion and the higher stability across different tests. 

The most relevant difference compared to the previous observation is related to the \%FMR at FMR=1\%. In particular, Table \ref{tab:crossdataset_table} shows that the best value between the verification systems enhanced through single 3DFR algorithms is obtained when NextFace is employed.
This outcome represents another proof of the complementarity between 3DFR algorithms.

\section{Key Insights and Guidelines}\label{sec:guidelines}

The trends observed in Section \ref{sec:results} allow us to draw some important considerations that may help technologists in system design and implementation of biometrics systems exploiting fusion-based 3D face reconstruction. First, the difference in performance between the best-performing non-parametric and parametric fusion methods suggests that no approach outperforms the other. 
This could be especially beneficial in cross-setting and cross-dataset contexts because it could be difficult to appropriately set the weights of the fusion module or train it with no data available in the specific scenario, as is typical in surveillance contexts. This drawback makes a non-parametric fusion like the average rule desirable in many real-world application scenarios.
Still, it is possible to observe results that are coherent with the previous literature related to facial biometrics when comparing the score-level fusion methods falling in the same category.

Other aspects that a technologist should take into account while selecting an appropriate score-level fusion method are:
\begin{itemize}
    \item The performance of the overall systems depends on both the underlying recognition systems included in the fusion and the capability of the score-level fusion method to exploit their complementarity.
    \item Based on the application context, the technologists should define whether they expect known or unknown data characteristics in terms of acquisition distance and camera and can obtain representative data for training a face recognition system.
    \item The technologist should define at which operational point the detection system will have to work based on the desired balance between false match and non-match errors.
    \item A 3DFR algorithm that provides the best results when enhancing a face verification system is not necessarily the best choice. Hence, the capability of the individual matcher to benefit from the specific 3DFR algorithm to extract useful information should be assessed.
    \item In the range between 1 and 4.2 meters, a higher use-case acquisition distance with respect to training data adversely impacts the overall performance more than a different (but still technologically similar) camera. Thus, when both conditions cannot be met, priority should be given to acquiring data at distances closer to those used in testing, rather than using images captured with the same camera but at different distances.
    \item There is no score-level fusion method that outperforms others in any operating conditions from both a global perspective and considering the single operational points. Therefore, the technologist should select a fusion rule according to the operating points required by the specific application context.
\end{itemize}

\section{Discussions and Conclusions}\label{sec:conclusions}
In this paper, we investigated the complementarity of four state-of-the-art 3D face reconstruction (3DFR) algorithms and the effectiveness of score-level fusion in exploiting their synergy to aid face recognition in challenging scenarios such as the surveillance one.
In particular, we assessed the complementarity from the linear correlation between the scores produced by different face verification systems, each based on the same deep-learning architecture but enhanced by a different 3DFR algorithm. Specifically, we evaluated several benchmark Siamese neural networks with backbones of varying computational complexity, including a verification system explicitly designed for face recognition in challenging surveillance scenarios.
Then, we explored various fusion methods, which can be roughly categorized into non-parametric fusion, fusion based on the weighted average, and fusion based on classification models for combining the systems based on the same backbone but enhanced through different 3D face reconstruction algorithms.

We also compared the investigated fusion methods using a unified experimental setup, evaluating their performance in both intra-setting and cross-setting scenarios. The experiments considered variations in acquisition distances and surveillance camera types and were conducted on a representative, publicly available dataset. To further assess the robustness of our approach, we extended the evaluation to a more challenging cross-dataset scenario, testing our method on an additional surveillance-related dataset that was not used for training the verification systems.

%Therefore, we compared such a set of fusion methods through a common experimental setup in intra-setting and cross-setting scenarios on a representative and publicly available dataset. These allowed us to study different application contexts: the intra-setting experiments simulated a context in which the operator knows the characteristics of the data (the acquisition camera and the distance from the captured subject) and has access to representative data for training the face recognition system; the cross-setting experiments simulated a more unfavourable application context in which the data characteristics are unknown or representative data cannot be obtained. Additionally, we assessed the potentiality of our approach on an even more challenging cross-dataset scenario, testing our approach on another surveillance-related dataset not employed for training the verification systems.

In general, the results obtained from an extensive set of experiments (\textit{i.e.}, 63 intra-setting, 420 cross-setting, and 63 cross-dataset) confirmed our initial hypothesis that a suitable fusion method can help improve performance and hence robustness to a face recognition system trained through a single 3DFR algorithm. Therefore, the primary outcome of our contribution is that the proposed approach can effectively exploit the complementarity between 3DFR algorithms. Based on the obtained results, we also provided a set of key insights and guidelines that could aid technologists and researchers in designing and implementing robust biometric systems. Note that the utility of the proposed fusion-based 3DFR has been specifically demonstrated for face verification in a challenging video surveillance setup, but our proposed methods may be applied to other tasks around face biometrics besides identity recognition.

%Moreover, our approach is not tied to the investigated systems and can, therefore, be included in other approaches based on deep learning for face verification beyond Siamese neural networks, for example, on state-of-the-art systems like PDT \cite{george2022prepended}.
%However, it is necessary to remark that score-level fusion is not able to offer improvements when the performance of all combined recognition systems is poor, as in the case of systems based on MobileNet.

According to what reported, deep-learning face recognition systems are able to extract distinct information from different state-of-the-art 3DFR algorithms, and the score-level fusion methods can exploit such a complementarity to make the inferences more robust, even concerning a challenging scenario such as a surveillance one. 
%The improvement can be observed even on probe images obtained from surveillance cameras and at distances different from the ones employed for training the data.
%Therefore, this approach could represent a valid tool for improving facial recognition in challenging scenarios like the one considered in this work.
However, it is essential to remark that our approach is not tied to the investigated systems and can, therefore, be included in other approaches based on deep learning for face verification beyond Siamese neural networks, for example, on state-of-the-art systems like PDT \cite{george2022prepended}.
Similarly, our approach is not necessarily an alternative but could be synergic with other well-known approaches to improve face recognition from pairs of high- and low-resolution images or perceptual quality of probe images, like resolution adaptation \cite{HU201746,grm2019face} and the introduction of different hard and soft biometrics \cite{maity2021multimodal,sosa2018facial}.
We did not include these strategies in this study to emphasize the potentialities of the only complementary information provided by distinct 3DFR models in face recognition. However, the synergy mentioned above should be investigated and included in systems aimed at aiding and speeding up the work of security enforcement officials and forensic practitioners. 
Furthermore, future studies should investigate alternative fusion approaches based on unsupervised machine learning or deep learning models. Our future work will also explore the proposed methods to improve facial biometrics in new applications not strictly related to identity recognition, \textit{e.g.}, e-health \cite{gomez2023} and e-learning \cite{daza2024}.

%Additionally, comparisons between score-level and decision-level fusion should be conducted to analyze their impact on improving generalization and the influence of lower-performing matchers on overall fusion performance.
%Additionally, the impact of the proposed approach on different pre-processing settings should be investigated. Furthermore, it should be investigated whether using models trained on data with different characteristics allows a higher ability to generalize and recognize individuals from combinations of acquisition distances and surveillance cameras not considered in the training set.

This study sheds some light on the differences between various fusion approaches in exploiting the complementarity information of distinct 3D face reconstruction algorithms in uncontrolled contexts. Therefore, we believe that the drafted path is promising and could contribute to aiding the development of multi-modal face recognition systems capable of recognizing never-seen-before individuals in challenging tasks such as identifying criminals and finding missing persons; and improving the robustness of new applications based on facial interaction beyond face verification.

\section*{Acknowledgments}
This work was supported by project SERICS (PE00000014) under the NRRP MUR program funded by the EU - NGEU, Cátedra ENIA UAM-VERIDAS en IA Responsable (NextGenerationEU PRTR TSI-100927-2023-2), and project BBforTAI (PID2021-127641OBI00 MICINN/FEDER). The work has been conducted within the sAIfer Lab and the ELLIS Unit Madrid.

\bibliographystyle{unsrt}  

\bibliography{references}

\end{document}